\documentclass{article}

\ifdefined\final
\else
\def\enablecomments{}
\fi

\ifdefined\draft
\def\enablecomments{}
\fi

\usepackage{xcolor}
\usepackage{soul}
\usepackage[normalem]{ulem}

\definecolor{LightGreen}{rgb}{0.80,1.00,0.80}
\definecolor{LightBlue}{rgb}{0.80,0.80,1.00}
\definecolor{LightRed}{rgb}{1.00,0.80,0.80}
\definecolor{LightPurple}{rgb}{1.00,0.80,1.00}

\soulregister{\method}{7}
\soulregister{\xspace}{7}
\soulregister{\emph}{7}
\soulregister{\cite}{7}

\ifdefined\enablecomments
  \DeclareRobustCommand%
  {\commentformat}[3]{\sethlcolor{#2}\textsf{\hl{#1: #3}}}
  \newcommand{\sm}    [1]{{\vspace{0.5em}\noindent\sethlcolor{LightGray}\hl{\textsf{#1}}}}
\else
  \newcommand{\commentformat}[3]{}
  \newcommand{\sm}    [1]{}
\fi

     \usepackage[nonatbib]{neurips_2020}

\usepackage[utf8]{inputenc} 
\usepackage[T1]{fontenc}    
\usepackage{hyperref}       
\usepackage{url}            
\usepackage{xtab}
\usepackage{booktabs}       
\usepackage{printlen}\uselengthunit{in}
\usepackage{amsfonts}       
\usepackage{nicefrac}       
\usepackage{microtype}      

\usepackage{caption}
\hypersetup{
    colorlinks=true,
    linkcolor=blue,
    filecolor=magenta,      
    urlcolor=cyan,
    citecolor=orange,
}
\usepackage[square, numbers]{natbib}
\usepackage{graphicx}
\begin{document}
\title{Efficient CNN-LSTM based Image Captioning using Neural Network Compression}

%

\author{%
  Harshit Rampal \\
  Carnegie Mellon University\\
  \texttt{hrampal@andrew.cmu.edu} \\
   \And
  Aman Mohanty \\
  Carnegie Mellon University\\
   \texttt{amanmoha@andrew.cmu.edu} \\
}

\maketitle


\begin{abstract}
Modern Neural Networks are eminent in achieving state of the art performance on tasks under Computer Vision, Natural Language Processing and related verticals. However, they are notorious for their voracious memory and compute appetite which further obstructs their deployment on resource limited edge devices. In order to achieve edge deployment, researchers have developed pruning and quantization algorithms to compress such networks without compromising their efficacy. Such compression algorithms are broadly experimented on standalone CNN and RNN architectures while in this work, we present an unconventional end to end compression pipeline of a CNN-LSTM based Image Captioning model. The model is trained using VGG16 or ResNet50 as an encoder and an LSTM decoder on the flickr8k dataset. We then examine the effects of different compression architectures on the model and design a compression architecture that achieves a 73.1\% reduction in model size, 71.3\% reduction in inference time and a 7.7\% increase in BLEU score as compared to its uncompressed counterpart.

\end{abstract}

\section{Introduction} \label{intro}

In recent years, deep neural networks have gained massive popularity for achieving state of the art results on tasks like classification, recognition and prediction. However, such complex networks have a large computational footprint that impedes their portability to low power mobile devices. Modern mobile devices have light and sleek form factors that further constraints their power and thermal capacity. For instance, a 32bit floating point addition operation consumes 0.9pJ, under 45nm CMOS technology. As summarized by \citet{han2015deep}, an on-chip SRAM cache access takes 5pJ, while an off-chip DRAM access takes 640pJ. Blatantly, complex neural networks will require an off-chip DRAM access which is way costlier.  From an energy perspective, a typical neural network having more than 1 billion connections, running at 30fps will take more than 19W for just accessing DRAM memory, which, is well beyond the power limit of a typical mobile device. 

Hence, there is a need to compress deep networks to enable their real-time applications on resource-limited devices. Recently, advanced pruning and quantization algorithms have gained momentum in compressing such networks without compromising their performance. Pruning helps in reducing the parameters that are less sensitive to a change in network's performance. On the other hand, quantization carries out the computations during a network’s work cycle in a lower bit precision. A synergy of these two methods enables faster inference times and efficient storage of large and dense neural networks.

Conventionally, researchers test their compression pipelines on standalone CNNs ranging from AlexNet\cite{krizhevsky2017imagenet} to MobileNet\cite{howard2017mobilenets} and even on RNNs and LSTMs. In this project, we followed an unconventional approach of developing and experimenting a compression pipeline on a novel use case i.e. a CNN-LSTM based image captioning model. As the name describes, an image captioning model generates captions or texts related to the contents of the image. The model is a sequence of an encoder (CNN) and a decoder (RNNs). The encoder extracts visual features from the image and feeds it to the decoder to generate captions. We use state-of-the-art CNNs like VGG16 and Resnet50 as encoders and build our decoder from scratch using LSTM. Such encoder-decoder models have huge number of parameters as they use both a CNN(e.g. VGG16 has 138M parameters) and an LSTM network (having 2.62M parameters), resulting in a  massive number of  network parameters. Deploying such huge networks on mobile devices is unfeasible due to the power, space and thermal constraints. Hence, an end-to-end compression of a captioning model is crucial to leverage its real time applications on mobile devices. Interestingly, a compressed version of such an image captioning model can be deployed on wearable electronics for assisting visually impaired people.  

We trained, validated and tested our compressed captioning model on flickr8k\footnote{\url{https://www.kaggle.com/adityajn105/flickr8k}} dataset. It consists of 8094 images each with 5 captions. The metric for evaluation was BLEU scores \cite{papineni2002bleu}, a commonly used method to evaluate a generated sequence with a reference sequence.

We employed magnitude based pruning to sparsify both the encoder and decoder parts of the network. We further implemented and experiment two types of quantization schemes, namely, post-training and quantization aware training.  Post training quantization was implemented on the encoder while both quantization schemes were experimented on the decoder. Section-\ref{results} portrays our findings on implementing different compression architectures  on the captioning model. Interestingly, some of them even outperformed the full-scale uncompressed model. On the basis of the reported results, we strongly advocate a particular compression architecture to compress the captioning model. The compressed model achieves impressive storage efficiency coupled with a respectable reduction in inference time.

\section{Related Work} \label{related}

Our literature review focuses on the different use cases of neural network compression and the different compression techniques. While there is an extensive repertoire of work on the different compression techniques, the use cases of such algorithms have not yet been widely explored.

Recently, Google launched Live Caption \cite{livecaption}, an on device captioning feature on its mobile devices making use of a RNN based model to caption audio sequences. For the purpose of deploying on mobile devices, Google used neural network pruning to reduce power consumption to 50\% while still being able to maintain the efficacy of the network \cite{shangguan2019optimizing}. \citet{kimvehicular} designed an object recognition system to recognize vehicles on the road using Faster RCNN. They were able to deploy the system on embedded devices by compressing the network using pruning and quantization. \citet{tan2019image} proposed pruning techniques on RNN for an image captioning model. While they were able to reach 95\% sparsity level without significant loss in performance, the proposed pruning techniques applied only to the RNN layers and not to the convolutional layers. The above work implemented and validated compression techniques on standalone CNNs, RNNs or LSTM architectures whereas we will be focusing on the effect of applying compression techniques on an end-to-end CNN-LSTM based image captioning pipeline.

There has been an immense advancement in research on neural network compression, especially on pruning and quantization in the last decade owing to deploying neural networks on resource-constrained environments. Among this large corpus of work, we review a select few which align best with our project. 

\citet{han2015learning} proposed a three step process to prune and fine tune a network to reduce a significant number of parameters without incurring massive performance loss. Along similar lines, \citet{zhu2017prune} validated the efficacy of magnitude based pruning by comparing pruned networks (large-sparse) and dense networks (small-dense) with similar memory footprints. \citet{li2016pruning} proposed a filter pruning method to remove filters having small effect on output accuracy and thereby reducing the convolution computation costs. The above methods follow an iterative approach to prune the networks after the network has been trained. On the other hand, \citet{lee2018snip} proposed a pruning algorithm to prune the network once at initialization prior to training.

\citet{han2015deep} proposed a compression pipeline by pruning, quantizing and encoding a neural network to significantly reduce a network's footprint with no loss in accuracy. After pruning the network, they reduce the number of bits required to represent the networks learned weights and limit the number of effective weights to store by having multiple connections share the same weight. They show that with pruning and quantization, a model can be compressed to 3\% of original size with no loss of accuracy whereas only pruning compresses the network to 8\% of its original size. \citet{jacob2018quantization} proposed a quantization scheme that relies only on integer arithmetic to approximate the floating-point computations in a neural network. They were able to achieve 4 times reduction of model size and improvement in inference efficiency in ARM NEON-based implementations.

\section{Method} \label{method}

\begin{figure}
    \centering
    \includegraphics[scale=0.5]{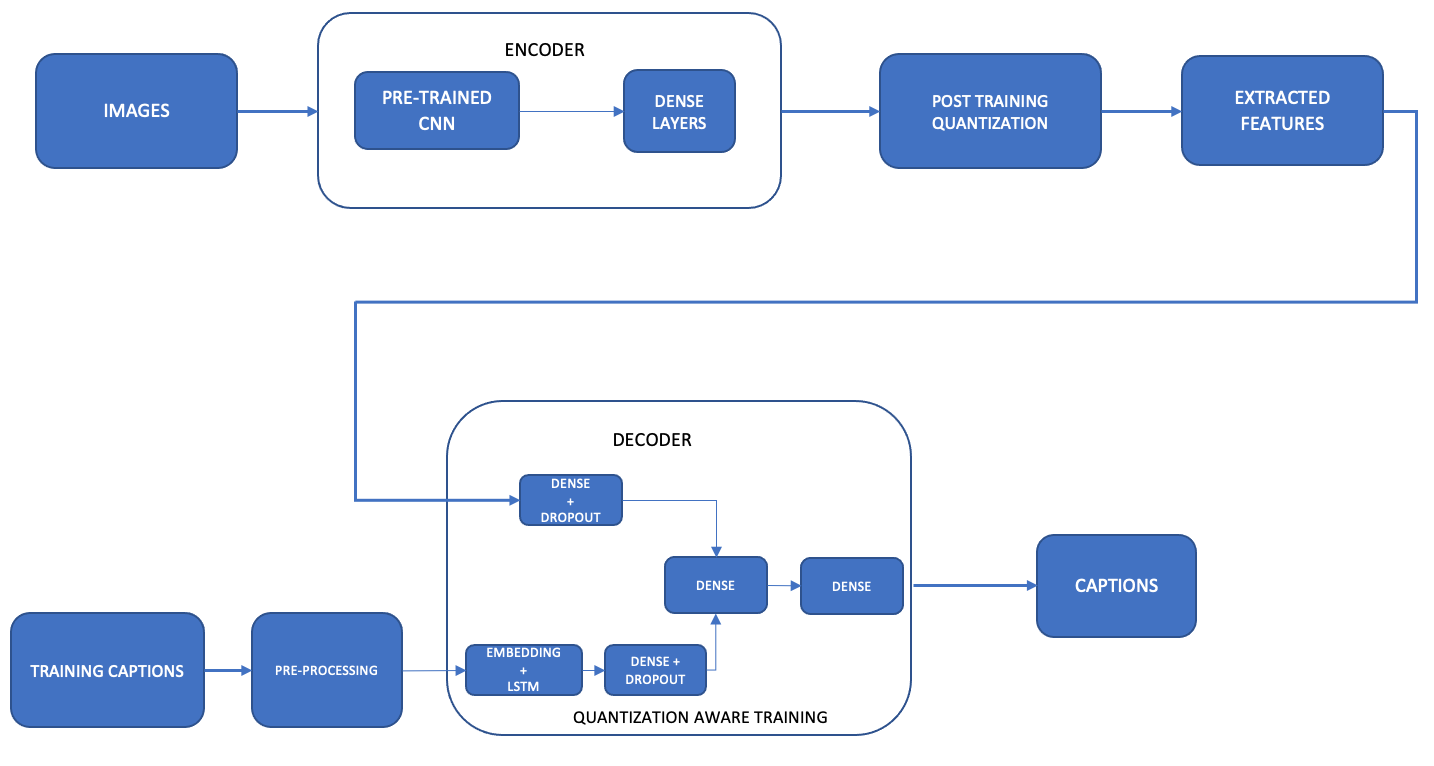}
    \caption{Training pipeline for the model with best compression architecture i.e., quantized encoder quantized decoder image captioning model.}
    \label{pipeline}
\end{figure}

An image captioning model consists of an encoder and a decoder. In our work, the encoder and the decoder have separate training procedures. We first train the encoder, obtain extracted features and then use those features to train the decoder. While deploying the network, we merge both the models. Figure-\ref{pipeline} demonstrates a complete training pipeline for the quantized encoder quantized decoder model.

\subsection{Encoder}

The role of the encoder is to extract meaningful features from an input image. We leveraged the power of established architectures, namely, VGG16 and ResNet50 for this task. 

Simultaneous working of our Encoder-Decoder model is a laborious task that requires a lot of RAM. So, in order to make the best use of our available computing resources, we adopted progressive storing. It is about separately storing the features extracted by the pruned and quantized pre-trained CNN model. These features are then fed to the decoder model for captioning. Hence, this strategy of separately loading the CNN features is more efficient as compared to the computationally demanding joint processing of the encoder-decoder model.   

\subsubsection{Pruning}
Pruning is a practice of zero-masking the weights that hold less importance. Here “importance” refers to their gradient with respect to the loss function, which, reflects their role in affecting the training accuracy of the network. There are various levels of pruning, generally, ranging from 50\% to 95\%. State of the art pruning algorithms are able to retain original levels of accuracy even after pruning 90\% of the weights. However, the upper bound of pruning sparsity is heavily dependent on the model architecture. Mainstream architectures ranging from AlexNet to MobileNet can be safely pruned in the range of 85\% to 95\%. It should be noted that pruning more than 95\% of the network parameters generally results in a significant dip in accuracy.

 The dense layers in the stated architectures were added to binary mask variables of the same corresponding shapes. The mask determines which weights will be made zero and the rest retain their value. The masks of the weights attaining a magnitude below than a threshold after $t_{0}$ number of epochs are set to zero. Here $t_{0}$ is a hyper-parameter.  Additionally, Pruning is defined for a sparsity range. The lower bound, $s_{i}$ is generally 0$\%$ and the upper bound, $s_{f}$, generally, varies from 50$\%$ to 95$\%$. The binary masks are updated every $\Delta$t steps as the network is gradually trained to reach the final sparsity level. Furthermore, zero-masked weights are not updated during backpropogation and this procedure of zero masking is followed for n pruning steps, until the selected layer attains the final sparsity level. \citet{zhu2017prune} contrived the following relation between $s_{i}$, $s_{f}$, n, t and $\Delta$t.

\begin{equation} \label{eu_eqn}
s_{t} = s_{f} + (s_{i} - s_{f})\left( 1 - \frac{t-t_{0}}{n\Delta t}\right)^3
\end{equation}

The above formulation conveys that the network undergoes extensive pruning during the first few epochs. After $t_{0}$ epochs, the pruning rate gradually decreases. Pruning is highly correlated with the learning rate as a small learning rate makes it difficult to recover from the accuracy loss induced by pruning while a large one will prune weights that hold significant importance to the networks performance. Lastly, when $s_{f}$ is achieved for the selected layer, binary masks are no longer updated. In order to prune the convolutional layers of these networks, they must be iteratively trained and retrained on ImageNet. This task was beyond the limit of our computing resources. So, we experimented pruning and quantizing the last dense layers of VGG16 and ResNet50.

\subsubsection{Quantization}
It is also desirable to quantize the network in conjunction with pruning for efficient use of storage. Quantization is about learning a reduced bit precision without compromising network performance and it is recognized as one of the most effective approaches to satisfy low memory requirements of resource constrained devices.  It is employed to store weights and subsequent matrix calculations during the forward, backward and update processes of a neural network in compact formats such as float-16, int-32 and even lower.  Such efficient use of storage enables fitting large models into an on-chip SRAM cache rather than an off-chip DRAM memory. 

Post Training quantization converts the weights from floating point representation to integer representation having 8 bits of precision. At inference time, activations are also converted to int8 format and further computations are done using the integer based weights. The following equation represents the 8bit post training quantization \cite{jacob2018quantization}. 
\begin{equation} 
\textrm{Value}_{float32} = \textrm{scale}\times(\textrm{Value}_{int8} - \textrm{Zeropoint}) \label{eu_eqn}
\end{equation} 
Per axis/per weights are represented by int8 two’s complement values in the range [-127,127] where zero point is equal to 0. Per tensor activations or per tensor inputs are represented by int8’s two complement values in the range [-128,127] have a zero point in the range of [-128,127]. Zero point follows the int8 representation of the zero point in the floating point precision. This ensures that 0 in float format is exactly representable by its quantized int8 value. The scale is a positive real number in floating point precision.

\subsection{Decoder}

The decoder produces captions for an image based on the extracted features of the image from the encoder. To train the decoder, we use the extracted features of the training images from the encoder model and the processed captions of those train images. We process the text using standard pre-processing procedures like converting characters to lower case, removing punctuation and digits. We then create the vocabulary followed by its tokenization. 

We design a multi-input decoder model to process the extracted features and the texts to produce captions. The \textit{feature extractor} part has a dense layer with 256 neurons and a dropout layer, the \textit{text extractor} part pre-process the training captions and is followed by an embedding layer and an LSTM layer of 256 neurons. The \textit{decoder} layer merges the outputs from the \textit{feature} and the \textit{text extractor} layers, followed by two dense layers, one with 256 neurons and the other with as many neurons as the vocabulary size. Training the decoder model as is requires a lot of memory. Due to our resource constraints, we train the decoder with progressive loading wherein we create a data generator function to provide a single sample of training data from the whole training set. We train the decoder sample by sample and that saves a lot of memory.

\subsubsection{Pruning}

We applied both pruning and quantization algorithms to compress the decoder model. We use the same pruning algorithm as used in the encoder. Each layer of the decoder model was pruned to achieve 50\% sparsity level. While pruning reduced the model size and produced BLEU scores comparable to the baseline model, the generated captions were not comprehensible and the model size reduction was not that significant. Hence, our best model doesn't incorporate decoder pruning. 

\subsubsection{Quantization}

For quantization, we employed both post-training quantization and quantization aware training to convert the weights of the decoder model to 8-bit precision. Post-training quantization, same as the one employed for encoder, didn't perform as expected and produced very low BLEU scores. Quantization aware training, on the other hand, produced BLEU scores better than the baseline model. Quantization aware training has the same objective as the post training quantization i.e., to reduce the precision to 8-bits.  The former is applied during the training phase of the network \cite{jacob2018quantization}.  Before performing quantization aware training, we initialized the model with the baseline model weights to achieve better test accuracy. Our best decoder model is based on quantization aware training.

\section{Results} \label{results}

As stated before, we test our compression pipeline on an image captioning model consisting of  pre-trained VGG16/ResNet50 encoder and a LSTM decoder trained from scratch. The image captioning model is trained on the flickr8k dataset using TensorFlow framework and Keras APIs. For compression, we make use of the Tensorflow model optimization library. For evaluation, we use BLEU scores, model size and inference time of the model on test set. 

Table-\ref{tab:resultsVGG} and Table-\ref{tab:resultsResNet} demonstrate the effect of different compression architectures on the image captioning model with VGG16 and ResNet50 as encoders, respectively. In these tables, we report the BLEU-1 score, size of the encoder and decoder models combined (in MB) and the inference time of the model on test set (in mins). Also, we report the \% change in the metrics of the compressed models from the metrics of the baseline model. 

Out of the many different compression architectures that we implemented and evaluated, we report the metrics for the following 8 configurations:\begin{itemize}
    \item baseline encoder and baseline decoder
    \item baseline encoder and 50\% pruned decoder
    \item baseline encoder and quantized decoder
    \item 50\% pruned encoder and baseline decoder
    \item quantized encoder and baseline decoder
    \item 50\% pruned encoder and 50\% pruned decoder
    \item quantized encoder and quantized decoder
    \item 50\% pruned and quantized encoder and quantized decoder
\end{itemize}

\textbf{Effect of Pruning:} A pruned model should achieve smaller model size than its baseline counterpart without hampering the model's performance. From  Table-\ref{tab:resultsVGG} and Table-\ref{tab:resultsResNet}, comparison of Model 1 to 2 and Model 4 to 6 demonstrates the effect of decoder pruning. The BLEU score for the pruned model is atleast at par with its corresponding baseline model and the pruned model achieves conceivable reduction in model size. Comparison of Model 1 to 4 shows that while encoder pruning achieves better reduction in model size than decoder pruning, it comes at the cost of reduced BLEU score. 

\textbf{Effect of Quantization:} Similarly, comparison of Model 1 to 3 and Model 5 to 7 demonstrates the effect of decoder quantization. In one case, the quantized model achieves comparable BLEU score while in the other it performs significantly better. While the quantized models achieve as good model size reduction as their pruned counterparts, they achieve significant reduction in inference time. Comparison of Model 1 to 5 and Model 3 to 7 shows that encoder quantized models perform at par with their corresponding baseline models. This shows that both encoder and decoder quantization is highly effective.

\begin{figure}[h]
    \centering
    \begin{tabular}{c c}
        \includegraphics[scale=0.25]{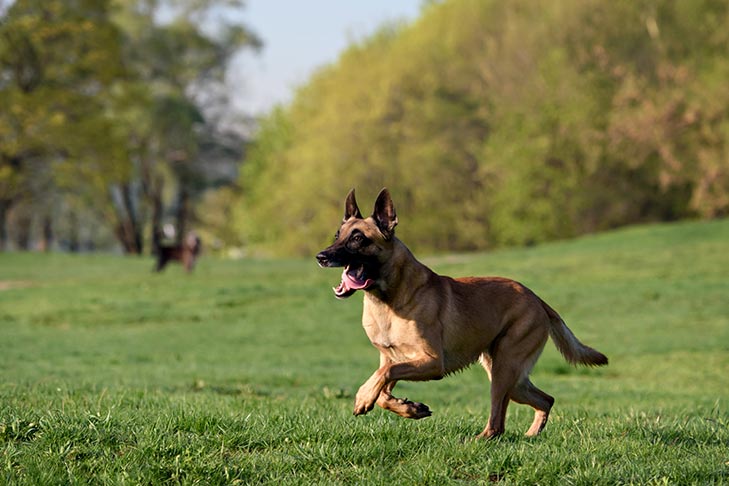} &
        \includegraphics[scale=0.297]{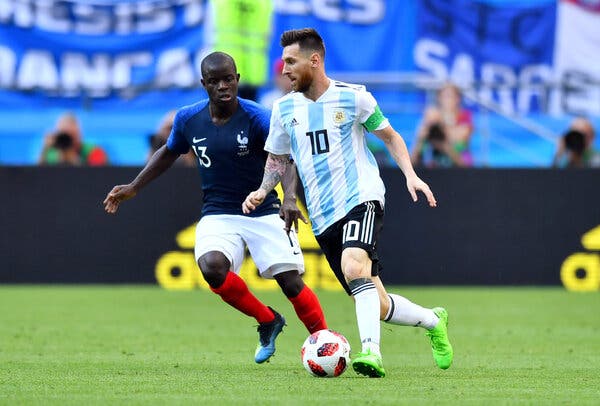} \\
        \small \textbf{Baseline}: two dogs are playing together in the grass & \small \textbf{Baseline}: two men are playing soccer on the grass \\
        \small \textbf{Best}: dog is running through the grass & \small \textbf{Best}: two men are playing soccer \\
        \includegraphics[scale=0.78]{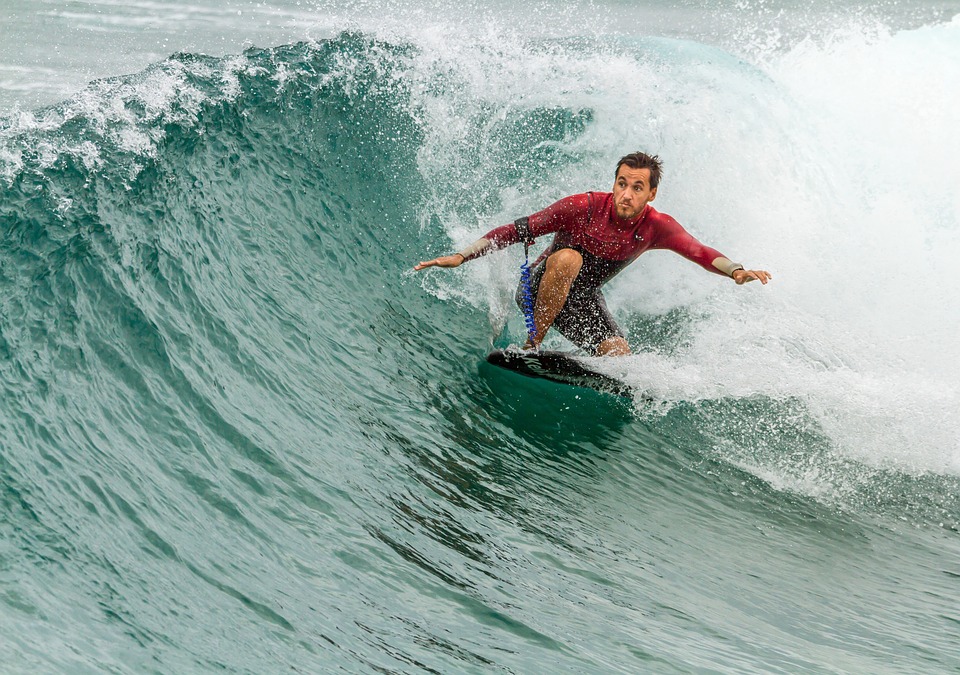} &
        \includegraphics[scale=0.126]{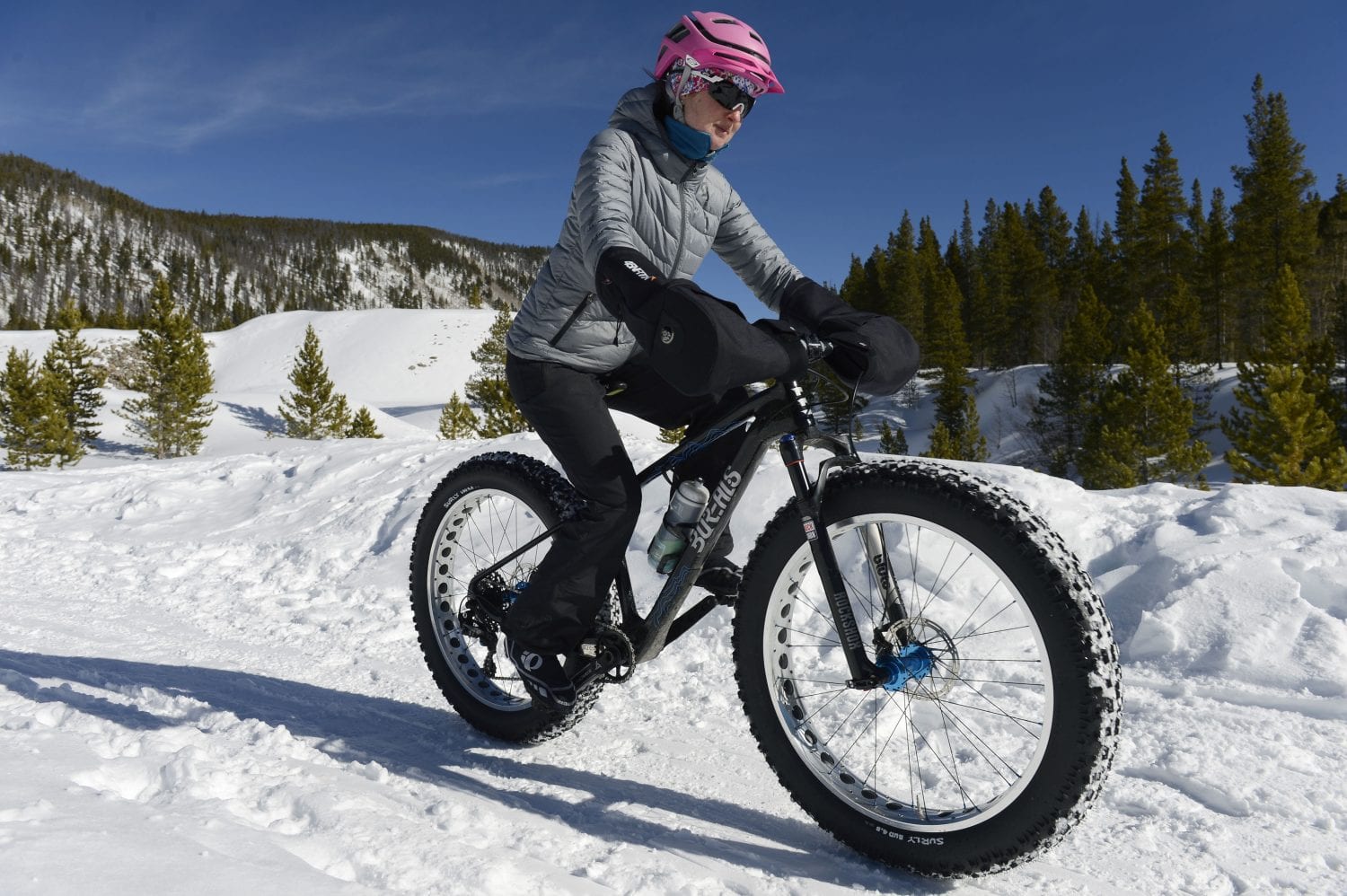} \\
        \small \textbf{Baseline}: man in red shirt is walking through the water & \small \textbf{Baseline}: man in red shirt is riding bike on dirt path \\
        \small \textbf{Best}: the man is in the water & \small \textbf{Best}: man in red shirt is riding bike on dirt path
    \end{tabular}
    
    \caption{Generated captions by the baseline and the best model, QVGG-QLSTM}
    \label{captions}
\end{figure}

We tried an interesting case of pruning and quantizing the encoder and only quantizing the decoder. From the tables, it can be observed that while this configuration achieves the lowest model size, the model performance is not comparable to the baseline. 
In Figure-\ref{captions}, we report the generated captions of four images taken from the internet [\href{https://s3.amazonaws.com/cdn-origin-etr.akc.org/wp-content/uploads/2018/06/05231748/belgian-malinois-running-through-field.jpg}{dog}, \href{https://static01.nyt.com/images/2020/09/25/sports/25soccer-nationalWEB1/merlin_177451008_91c7b66d-3c8a-4963-896e-54280f374b6d-articleLarge.jpg?quality=75&auto=webp&disable=upscale}{soccer}, \href{https://www.studyfinds.org/wp-content/uploads/2017/07/surfing-surfboard-man-surf-465216.jpeg}{surf}, \href{https://theknow.denverpost.com/wp-content/uploads/2018/11/TR25SNOWBIKE_AC18503x.jpg}{bike}]. 
Our implementation can be found at: \url{https://github.com/amanmohanty/idl-nncompress}.
\section{Evaluation} \label{evaluation}
Our attempt of pruning the decoder produced unanticipated results. A plausible reason for its under-performance can be attributed to the relatively simple architecture of the decoder. An attempt to prune the sole LSTM cell to 50$\%$ could have aggressively pruned its memory gate, thus affecting its efficacy of producing suitable captions. 

On the encoder side, pruning 50$\%$ of the dense layers in VGG16  and ResNet 50 led to appreciable results. Experiments with a higher sparsity level degraded the consistency of captions. The rationale for dip in accuracy lies in the limited representation of the terminal set of dense layers of both VGG16 and ResNet50. In order to implement pruning, one must iteratively train and re-train the network upto the desired sparsity level. Since, training and re-training on ImageNet dataset was beyond the limit of our resources, we switched to CIFAR-100 for pruning the terminal set of dense layers. CIFAR-100 is less variegated than ImageNet, so, the terminal layers must have learnt a limited set of features. Therefore, excessive pruning (above 50$\%$) of these layers dropped their efficacy in extracting better features.

The above reasoning provides an insight that the relative simplicity of the encoder and decoder makes the captioning model sensitive to a high level of pruning. Therefore, one may opine that the model performs best when pruning is avoided in both the encoder and decoder. Table-\ref{tab:resultsVGG} and Table-\ref{tab:resultsResNet} validate this opinion. 

We experimented only quantization aware training for decoder in place of coupling it with pruning. The primary reason is that the effect of pruning is diminished as quantization aware training is performed during training.

\begin{table}
    \centering
    \begin{tabular}{ c | c | c | c | c | c | c | c  }
    \toprule
    \multicolumn{1}{c|}{\textit{Sl No}}&
    \multicolumn{1}{c|}{\textit{Models}}&
    \multicolumn{2}{c|}{\textit{BLEU-1 score}}&
    \multicolumn{2}{c|}{\textit{Model size}}&
    \multicolumn{2}{c}{\textit{Inference time}} \\
    \cmidrule{3-8}
    & \textit{(Encoder-Decoder)} & \textit{score} & \textit{\% change} & \textit{MB} & \textit{\% reduction} & \textit{mins} & \textit{\% change} \\
    
    \midrule
    1 & Baseline-Baseline & 0.527 & - & 578.4 & - & 5.68 & - \\
    2 & Baseline-Prune & 0.534 & + 1.3 & 534.1 & 7.6 & 5.58 & - 1.7 \\
    3 & Baseline-Quantized & 0.521 & - 1.1 & 533.09 & 7.8 & 1.72 & - 69.7 \\
    4 & Prune-Baseline & 0.418 & - 20.6 & 74 & 87.2 & 5.79 & + 1.9 \\
    5 & Quantized-Baseline & 0.527 & 0 & 200.7 & 65.3 & 5.33 & - 6.2 \\
    6 & Prune-Prune & 0.418 & - 20.6 & 29.7 & 94.8 & 6.21 & + 9.3 \\
    7 & \textbf{Quantized-Quantized} & \textbf{0.568} & \textbf{+ 7.7} & \textbf{155.39} & \textbf{73.1} & \textbf{1.63} & \textbf{- 71.3} \\
    8 & PruneQuant-Quantized & 0.459 & - 12.9 & 22.99 & 96 & 1.74 & - 69.4 \\
    \bottomrule
    \end{tabular}
    \vspace{.05in}
    \caption{Evaluation of different compression architectures on the image captioning model with VGG16 as encoder trained on flickr8k dataset. Model sizes are reported in MB and inference times are reported in minutes taken per 2000 samples.}
    \label{tab:resultsVGG}
\end{table}

\begin{table}[h]
    \centering
    \begin{tabular}{ c | c | c | c | c | c | c | c  }
    \toprule
    \multicolumn{1}{c|}{\textit{Sl No}}&
    \multicolumn{1}{c|}{\textit{Models}}&
    \multicolumn{2}{c|}{\textit{BLEU-1 score}}&
    \multicolumn{2}{c|}{\textit{Model size}}&
    \multicolumn{2}{c}{\textit{Inference time}} \\
    \cmidrule{3-8}
    & \textit{(Encoder-Decoder)} & \textit{score} & \textit{\% change} & \textit{MB} & \textit{\% reduction} & \textit{mins} & \textit{\% change} \\
    
    \midrule
    1 & Baseline-Baseline & 0.51 & - & 150.1 & - & 5.91 & - \\
    2 & Baseline-Prune & 0.514 & + 0.8 & 110 & 26.72 & 5.7 & - 3.5 \\
    3 & Baseline-Quantized & 0.503 & - 1.3 & 109.1 & 27.31 & 1.65 & - 72.1 \\
    4 & Prune-Baseline & 0.418 & - 18.0 & 145.1 & 3.33 & 6.26 & + 5.9 \\
    5 & Quantized-Baseline & 0.546 & + 7.1 & 83.7 & 44.23 & 5.87 & - 0.7 \\
    6 & Prune-Prune & 0.418 & -18 & 105 & 30.04 & 4.79  & - 18.9 \\
    7 & \textbf{Quantized-Quantized} & \textbf{0.48} & \textbf{- 5.9} & \textbf{42.7} & \textbf{71.55} & \textbf{1.79} & \textbf{- 69.7} \\
    8 & PruneQuant-Quantized & 0.442 & -13.3 & 39.1 & 73.95 & 1.33 & - 77.5 \\
    \bottomrule
    \end{tabular}
    \vspace{.05in}
    \caption{Evaluation of different compression architectures on the image captioning model with ResNet50 as encoder trained on flickr8k dataset. Model sizes are reported in MB and inference times are reported in minutes taken per 2000 samples.}
    \label{tab:resultsResNet}
\end{table}

\section{Conclusion}
Among the corpus of different compression architectures that we implemented and evaluated on the image captioning model with VGG16 and ResNet50 as encoders, we select the VGG16-LSTM quantized encoder and quantized decoder (QVGG-QLSTM) as our best compressed model. The QVGG-QLSTM model performs better than the baseline model across all reported metrics. \textbf{It achieves a 7.7\% increase in BLEU score, 73.1\% reduction in model size and 71.3\% reduction in inference time per 2000 samples}.
In Figure-\ref{captions}, we report the generated captions of four images taken from the internet.

The captions reported are generated from the baseline model and our best model, QVGG-QLSTM. Clearly, the QVGG-QLSTM generates better captions for the 'dog' image than the baseline model. For the 'surf' and the 'soccer' image, both the models generate different but correct captions. For the 'bike' image both the models generate somewhat wrong but same captions. \textbf{This shows that the compressed model performs at par or better than the baseline model}.

\section{Future Work}
A couple of directions can be pursued to improve the present model. A run-of-the-mill method to boost performance of Deep Learning Neural Networks is to increase the size of the dataset. In this context, MS-COCO\cite{lin2014microsoft} is a promising choice. The dataset consists of 123,387 images, each with 5 captions. Another upgrade can be a more advanced pruning method. State of the art methods\cite{li2016pruning} not only prune the dense layers but the convolutional filters as well. This helps in achieving a higher level of sparsity without observing a significant dip in accuracy. Similarly, advanced quantization algorithms \cite{4bit} can be experimented to push the envelope of the present compression architecture.

\bibliography{references}







\end{document}